\documentclass[conference]{IEEEtran}
\IEEEoverridecommandlockouts
\usepackage{cite}
\usepackage{amsmath,amssymb,amsfonts,amsthm}
\usepackage{algorithmic}
\usepackage{graphicx}
\usepackage{textcomp}
\usepackage{xcolor}
\usepackage{url}
\usepackage{tikz}
\usepackage{pgfplots}
\usetikzlibrary{positioning}
\usepackage{caption}
\usepackage{subcaption}

\usepackage{booktabs}

\usepackage{colortbl}
\definecolor{Gray}{gray}{0.85}

\def\BibTeX{{\rm B\kern-.05em{\sc i\kern-.025em b}\kern-.08em
    T\kern-.1667em\lower.7ex\hbox{E}\kern-.125emX}}

\makeatletter
\newcommand{\linebreakand}{%
  \end{@IEEEauthorhalign}
  \hfill\mbox{}\par
  \mbox{}\hfill\begin{@IEEEauthorhalign}
}
\makeatother

\newcommand{\prob}[1]{\mathbb{P}\left[ #1 \right]}
\newcommand{\iid}{\overset{\textup{iid}}{\sim}}

\newcommand{\expe}[3][]{\mathbb{E}_{{#2}}\left[{#3}\right]}
\newcommand{\trsp}{\mathrm{T}}

\newtheorem{definition}{Definition}
\newtheorem{lemma}{Lemma}
\newtheorem{theorem}{Theorem}
\begin{document}

\title{P$^2$HCT: Plug-and-Play Hierarchical C2F Transformer for Multi-Scale Feature Fusion}

\author{\IEEEauthorblockN{Junyi Hu}
\IEEEauthorblockA{\textit{Department of Automation} \\
\textit{Tsinghua University}\\
Beijing, China \\
}
\and
\IEEEauthorblockN{Tian Bai}
\IEEEauthorblockA{\textit{Department of R\&D} \\
\textit{Linsulabs} \\
Chengdu, China \\
}
\and
\IEEEauthorblockN{Fengyi Wu}
\IEEEauthorblockA{\textit{School of Information and Communication Engineering} \\
\textit{University of Electronic Science and Technology of China}\\
Chengdu, China \\
}
\linebreakand 
\IEEEauthorblockN{Zhenming Peng}
\IEEEauthorblockA{\textit{School of Information and Communication Engineering} \\
\textit{University of Electronic Science and Technology of China}\\
Chengdu, China \\
}
\and
\IEEEauthorblockN{Yi Zhang}
\IEEEauthorblockA{\textit{Department of Automation} \\
\textit{Tsinghua University}\\
Beijing, China \\
}
}

\maketitle

\begin{abstract}
Feature fusion plays a pivotal role in achieving high performance in vision models, yet existing attention-based fusion techniques often suffer from substantial computational overhead and implementation complexity, particularly in resource-constrained settings. To address these limitations, we introduce the Plug-and-Play Hierarchical C2F Transformer (P$^2$HCT), a lightweight module that combines coarse-to-fine token selection with shared attention parameters to preserve spatial details while reducing inference cost. P$^2$HCT is trainable using coarse attention alone and can be seamlessly activated at inference to enhance accuracy without retraining. Integrated into real-time detectors such as YOLOv11-N/S/M, P$^2$HCT achieves mAP gains of 0.9\%, 0.5\%, and 0.4\% on MS COCO with minimal latency increase. Similarly, embedding P$^2$HCT into ResNet-18/50/101 backbones improves ImageNet top-1 accuracy by 6.5\%, 1.7\%, and 1.0\%, respectively. These results underscore P$^2$HCT’s effectiveness as a hardware-friendly and general-purpose enhancement for both detection and classification tasks. Our code is available at \url{https://github.com/inlmouse/P2HCT}.
\end{abstract}

\begin{IEEEkeywords}
Hierarchical C2F Transformer, Multi-Scale Feature Fusion, Coarse-to-Fine Attention
\end{IEEEkeywords}

\section{Introduction}
\label{sec:intro}

Feature fusion---integrating information across multiple layers or branches---serves as a cornerstone of contemporary visual architectures. Transformer-based models \cite{vaswani2017attention} have illustrated that \textbf{content-adaptive attention mechanisms} consistently surpass fixed convolutional schemes by representing image features as dynamic token sequences \cite{dosovitskiy2021an}. The resultant shift toward tokenization has made Transformer-based backbones and heads the dominant choice in state-of-the-art systems for visual recognition, detection, and segmentation.

\begin{figure*}[t]
  \centering
    \centering
\begin{tikzpicture}[scale=0.5]
  \begin{axis}[
      xlabel={FLOPs (G)},
      ylabel={MS COCO mAP\textsuperscript{test-dev}\textsubscript{50:95} (\%)},
      label style={font=\LARGE},
      grid=both,
      width=14cm, height=12cm,
      xmin=35, xmax=215,
      ymin=37, ymax=52,
      legend style={at={(0.98,0.5)}, anchor=east},
      legend style={font=\large},  
      legend style={
          fill=white,            
          fill opacity=0.5,      
          draw=black,            
          draw opacity=0.8,      
          text opacity=1,        
        }
    ]
    
    \addplot[mark=star,  mark size=3pt, thick, red, line width=2pt] coordinates {
      (42.5,46.3) (71.9,47.9) (138.5,50.9)
    };
    \addlegendentry{P$^2$HCT-DET (Ours)}

    \addplot[mark=o, thick, black] coordinates {
      (79.5,45.8) (113.3,46.9)
    };
    \addlegendentry{G-FPN}

    \addplot[mark=+, thick, green!60!black] coordinates {
      (136.4,38.4) (203.2,40.2)
    };
    \addlegendentry{AFPN}

    \addplot[mark=x, thick, brown] coordinates {
      (135.9,40.4) (202.7,42.4)
    };
    \addlegendentry{AC-FPN}

    \addplot[mark=square, thick, cyan] coordinates {
      (97.4,38.0) (129.4,40.0)
    };
    \addlegendentry{Retina+SA}

    \addplot[mark=triangle, thick, orange] coordinates {
      (120.4,39.8) (209.0,42.3)
    };
    \addlegendentry{A2-FPN}

    \node[anchor=west, font=\large, red] at (axis cs:36.5,47) {R18-P$^2$HCT-N};
    \node[anchor=south, font=\large, red] at (axis cs:123,48.3) {R50-P$^2$HCT-N};
    \node[anchor=east, font=\large, red] at (axis cs:200.5,51.2) {R101-P$^2$HCT-S};
    \node[anchor=north, font=\large, green!60!black] at (axis cs:136.4,38.4) {R-50-AFPN};
    \node[anchor=north east, font=\large, green!60!black] at (axis cs:210.2,39.7) {R-101-AFPN};
    \node[anchor=south, font=\large, brown] at (axis cs:120.9,40.5) {R-50-AC-FPN};
    \node[anchor=south east, font=\large, brown] at (axis cs:190,41.8) {R-101-AC-FPN};
    \node[anchor=north east, font=\large, cyan] at (axis cs:120,38.0) {R-50-Retina+SA};
    \node[anchor=west, font=\large, cyan] at (axis cs:129.4,39.8) {R-101-Retina+SA};
    \node[anchor=east, font=\large, orange] at (axis cs:120.4,39.8) {R-50-A²-FPN-Lite};
    \node[anchor=north east, font=\large, orange] at (axis cs:212.0,41.8) {R-101-A²-FPN};
    \node[anchor=north, font=\large] at (axis cs:65.5,45.5) {G-FPN-D7};
    \node[anchor=south, font=\large] at (axis cs:140,46.8) {G-FPN-D11};
  \end{axis}
\end{tikzpicture}
    \centering
    \begin{tikzpicture}[scale=0.5]
  \begin{axis}[
      xlabel={Latency (ms)},
      ylabel={MS COCO mAP\textsuperscript{val}\textsubscript{50:95} (\%)},
      label style={font=\LARGE},
      tick label style={font=\LARGE},
      grid=both,
      width=14cm, height=12cm,
      xmin=1, xmax=6,
      ymin=36, ymax=53,
      legend pos=south east,
      legend style={font=\large},
      legend style={
          fill=white,            
          fill opacity=0.5,      
          draw=black,            
          draw opacity=0.8,      
          text opacity=1,        
        }
    ]
    \addplot[mark=*, red, thick, line width=2pt, mark size =2.5] coordinates {
      (1.24,40.3)   
      (2.49,47.4)   
      (4.3,52.1)   
    };
    \addlegendentry{YOLOv11‑P$^2$HCT (Ours)}

    \addplot[mark=o, blue, thick] coordinates {
      (2.69,37.0)   
      (3.42,44.3)   
      (5.63,49.1)   
    };
    \addlegendentry{YOLOv6-3.0}

    \addplot[mark=square, orange, thick, dashed] coordinates {
      (1.77,37.4)    
      (2.33,45.0)   
      (5.09,50.3)   
    };
    \addlegendentry{YOLOv8}


    \addplot[mark=diamond, pink, thick, dashed] coordinates {
      (1.84,38.5)    
      (2.49,46.3)   
      (4.74,51.1)   
    };
    \addlegendentry{YOLOv10}

    \addplot[mark=o, black, thick, dashed] coordinates {
      (1.5,39.4)    
      (2.51,46.9)   
      (4.7,51.5)   
    };
    \addlegendentry{YOLOv11}



    \addplot+[
      only marks,               
      mark=square*,                   
      mark options={            
        fill=purple,               
        draw=purple,             
        scale=1.25              
      }
    ] coordinates {
      (2.3,42.6)                  
    };
    \addlegendentry{YOLOv12-P$^2$HCT-N (Ours)}
    
    \addplot+[
      only marks,               
      mark=square*,                   
      mark options={            
        fill=yellow!90!black,               
        draw=yellow!90!black,             
        scale=1               
      }
    ] coordinates {
      (1.64,40.6)                  
    };
    \addlegendentry{YOLOv12-N}
    \addplot+[
      only marks,               
      mark=square*,                   
      mark options={            
        fill=green,               
        draw=green,             
        scale=1               
      }
    ] coordinates {
      (1.35,39.1)                  
    };
    \addlegendentry{EfficientDet-D1}
  \end{axis}
\end{tikzpicture}
  \caption{Comparative performance of P$^2$HCT. (Left) Against attention-based FPN methods on the MS COCO test-dev set, P$^2$HCT demonstrates substantial accuracy gains with a streamlined, low-complexity design. (Right) When integrated into real-time object detection frameworks on the MS COCO validation set, its high-efficiency attention kernel enables state-of-the-art results with reduced latency.}
    \label{fig:yolo-tradeoffs}
\end{figure*}

Despite their representational power, existing attention-based fusion schemes (e.g., AttnFPN \cite{min2022attentional}, Deformable DETR \cite{zhu2020deformable}) incur \textbf{high FLOPs, irregular memory access, and engineering complexity}. Sparse approximations \cite{choromanski2020rethinking} often break hardware-friendly dataflows, and even sophisticated libraries (e.g., FlashAttention \cite{dao2022flashattention}) still operate on dense token grids, failing to fully resolve the efficiency bottleneck.

This motivates the question:
\emph{Can we design a plug-and-play Transformer module for visual feature fusion that (i) lightweight but state-of-the-art performance, (ii) drastically reduces complexity and (iii) preserves spatial fidelity?}

To address these challenges, we propose the \textbf{PnP Hierarchical C2F Transformer (P$^2$HCT)}. P$^2$HCT is a hierarchical fusion architecture designed for efficient cross-scale feature integration that serves as an effective, \textbf{plug-and-play} alternative to standard modules like FPN in detection heads. Our method consistently improves performance across model scales, achieving at least \textbf{0.4\% mAP gains on COCO}~\cite{lin2014microsoft} and enhanced top-1 accuracy on ImageNet~\cite{deng2009imagenet}, with minimal computational overhead.

Our key technical contributions are:

\begin{enumerate}
    \item We introduce a novel \textbf{PnP Hierarchical C2F Attention} that efficiently fuses coarse global context with sparse fine local details. This two-stage mechanism significantly reduces complexity from the conventional $\mathcal{O}(N^2)$ to a combination of $\mathcal{O}(\frac{1}{4}N^2)$ for the cross-layer coarse stage and $\mathcal{O}(4Nk)$ for the sparse fine stage, drastically improving computational efficiency.

    \item We develop a theoretically justified \textbf{parameter sharing scheme} that enables training solely with the computationally cheaper coarse branch. This design preserves the flexibility to activate the full fine attention during inference, allowing adaptable accuracy-efficiency trade-offs (proof provided in the supplemental material).

    \item The resulting architecture is extremely lightweight, matching the parameter footprint of a single $4{\times}4$ convolution, making it highly suitable for real-world deployment.
\end{enumerate}

\section{Related Work}
\label{sec:related}

\subsection{Hierarchical and Cross-Scale Feature Fusion Architectures}
Multi-scale feature fusion has long been essential in dense vision tasks. The foundational method is the \textbf{Feature Pyramid Network (FPN)} \cite{lin2017feature}, which establishes a top-down pathway with lateral connections to merge high-level semantic information into lower-level feature maps.

Modern \textbf{CNN-based} variants have significantly enhanced this design, often incorporating attention or sophisticated connection schemes. Examples include: BiFPN \cite{tan2020efficientdet} and G-FPN \cite{tan2021giraffedet} utilizing bidirectional paths; and attention-based fusion methods like A\textsuperscript{2}-FPN \cite{hu2021a2} and AttnFPN \cite{min2022attentional} embedding attention modules to adaptively weigh feature contributions. These methods, while effective, often rely on dense operations or full self-attention, leading to high computational costs and complex engineering.

More sophisticated cross-hierarchy attention methods (e.g., CFPT \cite{du2025cross}) achieve feature interaction via heavy attention blocks, incurring substantial computational and implementation overhead. In stark contrast, our \textbf{P$^2$HCT} module adopts a simpler, yet highly effective, cross-layer design: deeper features directly guide adjacent shallower ones via a single \textbf{lightweight attention block}. This simplicity makes it extremely efficient and allows it to seamlessly replace or integrate into common CNN-based task heads like FPN.

\subsection{Sparse and Efficient Attention for Vision}

The search for efficient alternatives to dense attention is a major research direction. While specialized kernels optimize the speed of dense computations, they do not reduce the inherent quadratic complexity, $\mathcal{O}(N^2)$, of the attention mechanism itself.

A parallel effort lies in \textbf{Dynamic Token Pruning} methods \cite{rao2021dynamicvit, liu2022adaptive, bonnaerens2023learned}, which aim to select the most informative subset of tokens to improve the computation-per-performance trade-off. However, these dynamic methods typically: (1) involve non-differentiable or multi-stage pruning steps, and (2) require complex training setups, such as auxiliary losses or specialized fine-tuning, to ensure stable convergence.

Our \textbf{P$^2$HCT} module addresses these limitations by leveraging a \textbf{top-$k$ selection} mechanism to dynamically focus on informative regions, but crucial to its stability is the \textbf{parameter sharing scheme}. This allows for stable training using only the computationally cheap coarse path, while the sparse fine attention path can be flexibly activated during inference for an accuracy boost, achieving dynamic efficiency \textbf{without training overhead or complex auxiliary losses.}

\section{Methodology}
\label{sec:method}

\subsection{Module Architecture of P$^2$HCT}

\begin{figure*}[t]
	\centering	\includegraphics[width=0.9\linewidth]{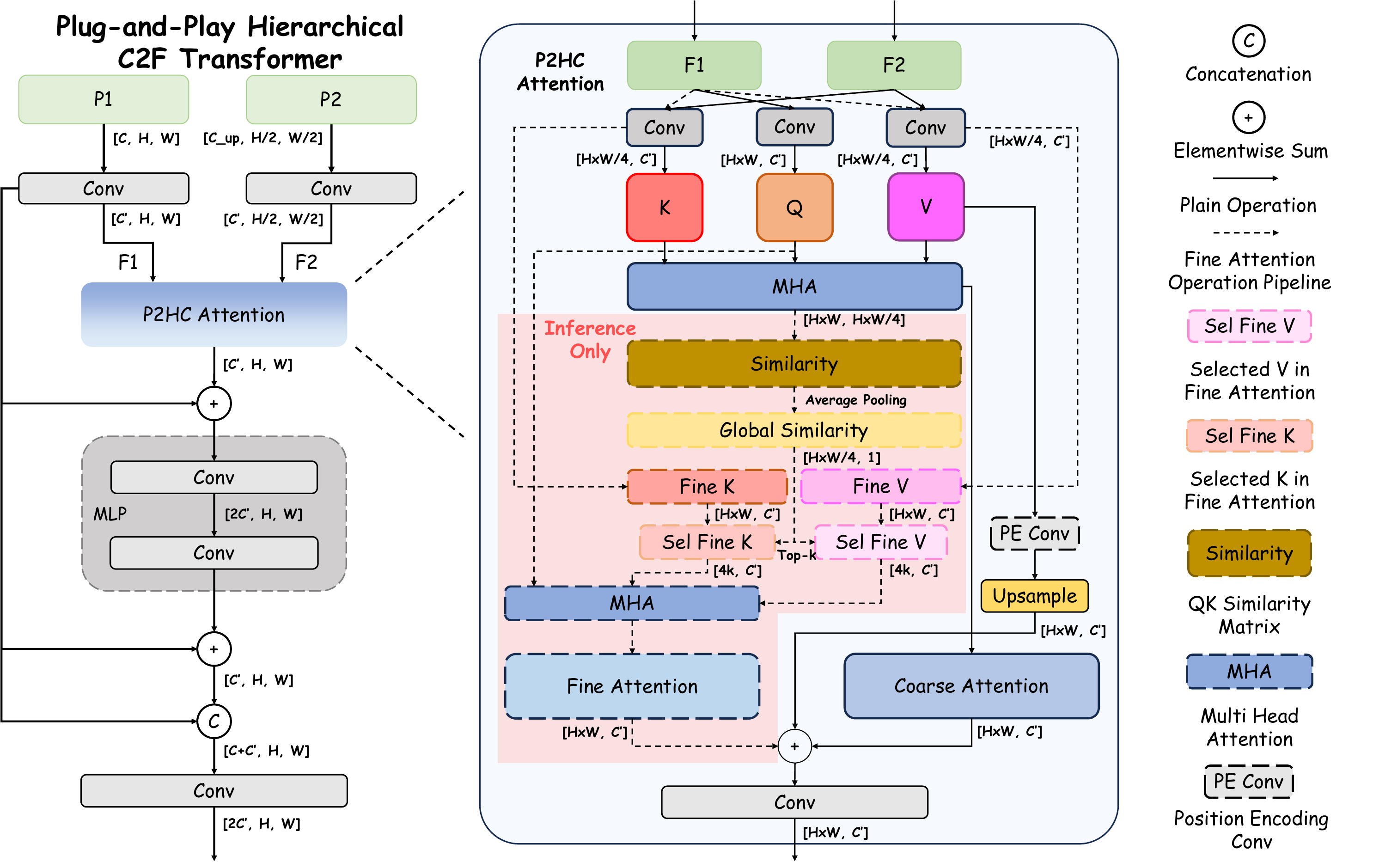}
	\caption{\textbf{Overview} of the PnP Hierarchical C2F Transformer. \textbf{Left:} P$^2$HCT takes two adjacent feature maps as inputs and replaces standard attention with our PnP Hierarchical C2F Attention (P2HCA) block. Drawing on the hardware friendliness and efficient design of EfficientFormer \cite{li2022efficientformer}, all Linear and LayerNorm layers are substituted with 1\(\times\)1 Convs followed by BatchNorm. Additionally, the concluding feature concatenation leverages insights from well-established architectures such as C3K2 \cite{wang2024yolov9}, renowned for their robust feature aggregation capabilities as demonstrated in prior empirical studies. \textbf{Right:} P2HCA first applies a cross-layer coarse attention, then selects the top-\(k\) fine tokens to perform sparse refined attention and sums both outputs.}
\label{fig:P$^2$HCT-Arch}
\end{figure*}
\begin{figure}[t]
  \centering
  \begin{subfigure}[b]{0.49\textwidth}
    \centering
    \includegraphics[width=0.87\textwidth]{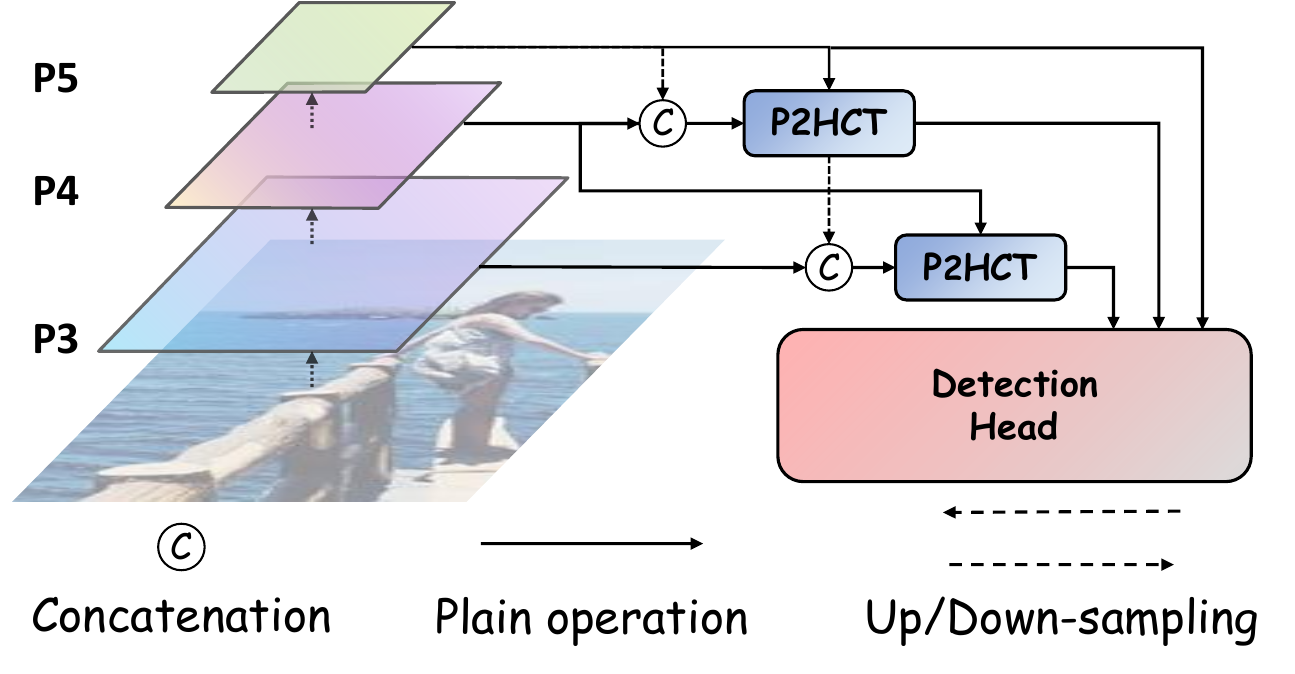}
    \label{fig:P2HCT-DET}
  \end{subfigure}
  \\
  \begin{subfigure}[b]{0.49\textwidth}
    \centering
    \includegraphics[width=0.87\textwidth]{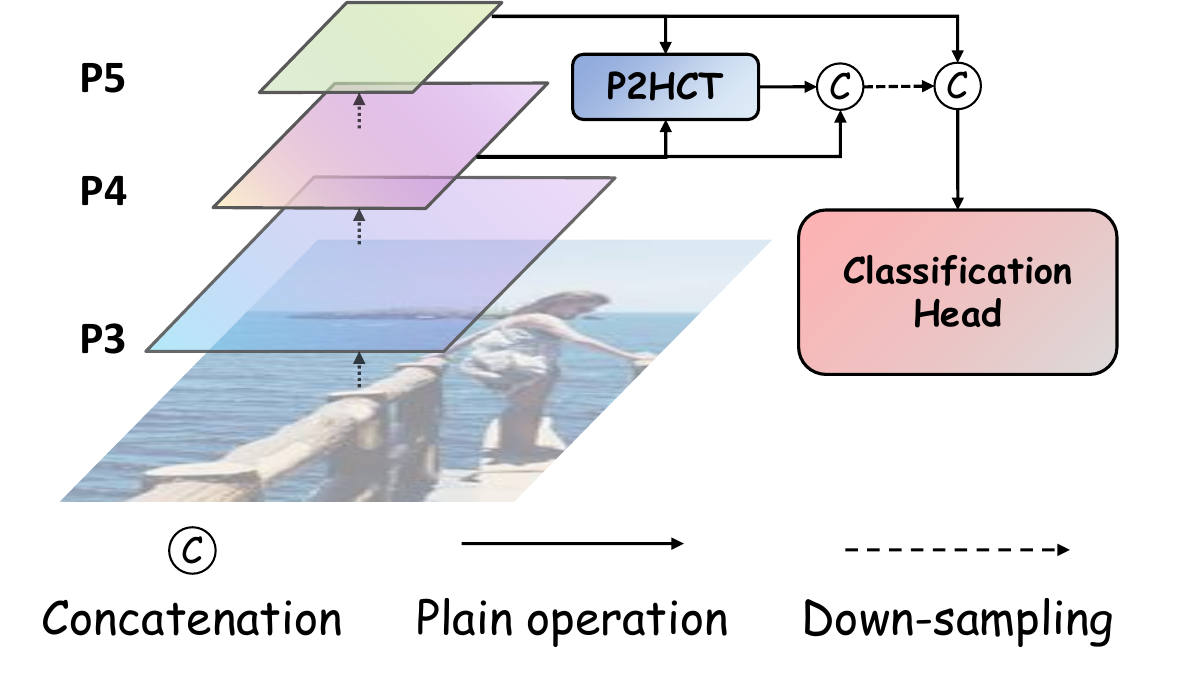}
    \label{fig:P2HCT-CLS}
  \end{subfigure}
  \caption{Frameworks of downstream tasks for (Top) detection and (Bottom) classification with our P$^2$HCT.}
  \label{fig:downstreamtasks}
\end{figure}

Building on the insight that high-level representations encode semantic abstractions while low-level features preserve spatial details, we propose the \textbf{PnP Hierarchical C2F Transformer (P$^2$HCT)}. Unlike standard self-attention which computes $\mathcal{O}(N^2)$ interactions within a single scale, P$^2$HCT adopts a cross-layer design (Figure~\ref{fig:P$^2$HCT-Arch}): it utilizes semantically rich upper-level features as Keys/Values ($\mathbf{K}, \mathbf{V}$) to guide Queries ($\mathbf{Q}$) from lower-level maps. To ensure spatial fidelity, we incorporate a Convolutional Positional Encoding (CPE) via a $7{\times}7$ depthwise convolution~\cite{chu2021conditional}. Crucially, we employ a \textbf{parameter sharing scheme} between coarse and fine stages, enabling efficient training using only the coarse branch while allowing flexible inference activation.

\noindent\textbf{Hierarchical Coarse-to-Fine Formulation.}
Formally, let $X \in \mathbb{R}^{C' \times H \times W}$ denote the low-level feature map and $U \in \mathbb{R}^{C' \times \frac{H}{2} \times \frac{W}{2}}$ the adjacent high-level map. We first project $X$ to Queries and $U$ to Keys/Values using $1{\times}1$ convolutions. The \emph{coarse attention} is computed by interacting full-resolution queries with downsampled keys:
\begin{equation}
    \mathbf{O}_{\mathrm{coarse}} = \text{Softmax}\left(\frac{\mathbf{Q}\mathbf{K}^\top}{\sqrt{d_k}}\right)\mathbf{V},
\end{equation}
where $\mathbf{Q}=\phi_q(X)$ and $\{\mathbf{K},\mathbf{V}\}=\phi_{k,v}(U)$. Since $|\mathbf{K}| = N/4$, this reduces complexity to $\frac{1}{4}\mathcal{O}(N^2)$. To capture fine-grained details, we calculate a global importance score $\mathbf{s} \in \mathbb{R}^{N/4}$ by averaging attention weights across queries and select the top-$k$ indices $\mathcal{I}$. These indices are mapped to the finer grid to retrieve sparse keys/values from $X$:
\begin{equation}
    \mathbf{K}_{\mathrm{fine}}^{\mathrm{sel}}, \mathbf{V}_{\mathrm{fine}}^{\mathrm{sel}} = \text{Gather}(\phi_k(X), \mathcal{I}), \quad \text{Gather}(\phi_v(X), \mathcal{I}).
\end{equation}
\emph{Refined attention} is then computed strictly within these selected regions, incurring only $\mathcal{O}(4Nk)$ complexity:
\begin{equation}
    \mathbf{O}_{\mathrm{fine}} = \text{Softmax}\left(\frac{\mathbf{Q} (\mathbf{K}_{\mathrm{fine}}^{\mathrm{sel}})^\top}{\sqrt{d_k}}\right)\mathbf{V}_{\mathrm{fine}}^{\mathrm{sel}}.
\end{equation}
Finally, the output aggregates the coarse context, fine details, and the CPE-enhanced residue via a linear projection: $\mathbf{O} = \phi_o(\mathbf{O}_{\mathrm{coarse}} + \mathbf{O}_{\mathrm{fine}} + \text{CPE}(\mathbf{V}))$.

\noindent\textbf{Complexity Analysis.}
P$^2$HCT achieves an overall complexity of $\tfrac14\mathcal{O}(N^2) + \mathcal{O}(4Nk)$. The architecture is designed to be extremely lightweight, consisting primarily of ten $1{\times}1$ convolutional layers. The total parameter count approximates:
\begin{equation}
    P_{\text{total}} \approx 10C'^2 + (C_{\mathrm{up}} + 3C)C' \approx 16C'^2,
\end{equation}
where $C'$ is the embedding dimension. Empirically, this footprint matches a single $4{\times}4$ convolution, making P$^2$HCT a highly efficient, plug-and-play solution suitable for modern hardware accelerators.

\subsection{Structure of P$^2$HCT-DET for Detection}
The P$^2$HCT-DET structure, illustrated at the top of Figure \ref{fig:downstreamtasks}, modifies the traditional FPN by integrating P$^2$HCT modules as plugins. In this architecture, the original convolutional layers within the FPN are replaced with P$^2$HCT modules, which process concatenated features from pyramid levels P3, P4, and P5. These modules adapt to the input through upsampling and downsampling operations, ensuring compatibility with varying feature resolutions. The refined features are subsequently fed into the detection head, enabling robust object detection across multiple scales. This plug-and-play design seamlessly enhances the FPN framework by leveraging P$^2$HCT’s cross-attention capabilities.

\subsection{Structure of P$^2$HCT-CLS for Classification}

The P$^2$HCT-CLS structure, depicted in the bottom of Figure \ref{fig:downstreamtasks}, employs P$^2$HCT to fuse high-level features from P4 and P5 layers, acting as a plugin positioned after the backbone network. By integrating these semantically rich features through the P$^2$HCT module, the architecture captures comprehensive contextual information, which is then passed to the classification head. This integration enhances classification accuracy without necessitating extensive modifications to the existing backbone, demonstrating P$^2$HCT’s versatility and effectiveness for feature fusion enhancement.


\section{Experiment}
\subsection{Implementation Detail}
\textbf{Detection Task:} We validate our method on the MS COCO 2017 dataset \cite{lin2014microsoft}. The detection experiments are split into two parts. The first part uses ResNet-18, ResNet-50, and ResNet-101 as backbone networks, integrating the P$^2$HCT-DET structure, and compares its performance against other attention-based FPN methods with the same backbones. The second part focuses on real-time detection, employing YOLOv11 variants (Nano, Small, Medium), where the head is replaced with the P$^2$HCT architecture, benchmarked against state-of-the-art real-time detection frameworks. All models are trained for 600 epochs using the SGD optimizer with an initial learning rate of 0.01, consistent with YOLOv11 \cite{jocher2024yolov11}.

\label{subsec:setup}
\begin{table}[t]
\centering
\setlength{\tabcolsep}{1mm}
\small
\begin{tabular}{lccccccc}
\toprule
\textbf{{Method}} & \textbf{F. (G)} & \textbf{\#P. (M)} & \textbf{$\text{AP}$ } &\textbf{$\text{AP}_{50} $ }  &\textbf{$\text{AP}_{75}$ } \\
\midrule
R-50-AttnFPN & 136.4*  & 46.9* & 38.4 & 61.1 & 41.9  \\
R-50-$A^2$-FPN-Lite & 120.4*  & 44.5* & 39.8 & 62.3 & 43.4  \\
R-50-AC-FPN & 135.9*  & 44.7* & 40.4 & 63.0 & 44.0  \\
G-FPN-D7 & 79.5  & 4.54* & 45.8 & - & -  \\
R-50-AFPN & 90.0  & 50.2 & 39.0 & 57.6 & 42.1  \\
\rowcolor{Gray}
\textbf{R-18-P$^2$HCT-N(Ours)}& 42.5  & 16.4 & \textbf{46.3} & \textbf{63.7} & \textbf{49.6}  \\
\rowcolor{Gray}
\textbf{R-50-P$^2$HCT-N(Ours)}& 71.9  & 26.0 & \textbf{47.9} & \textbf{65.3} & \textbf{51.8}  \\
\midrule
R-101-Retina+SA & 129.44  & 58.5 & 40.0 & 60.0 & 42.9  \\
R-101-AttnFPN & 203.2*  & 64.7* & 40.2 & 62.5 & 43.6  \\
R-101-$A^2$-FPN & 209.0*  & 62.5* & 42.8 & 65.2 & 47.0  \\
R-101-AC-FPN & 202.7*  & 62.5* & 42.4 & 65.1 & 46.2  \\
G-FPN-D11 & \textbf{113.3}  & \textbf{7.2}* & 46.9 & - & -  \\
\rowcolor{Gray}
\textbf{R-101-P$^2$HCT-S(Ours)}& 138.5  & 49.9 & \textbf{50.9} & \textbf{68.1} & \textbf{55.7}  \\
\midrule
\end{tabular}
\caption{Comparison with popular state-of-the-art Attention FPN based object detectors on MS COCO 2017 test-dev set. All results are obtained using $640\times640$ inputs. $\text{AP}$s are in the unit of \%. ~*Indicates the data is not publicly available and is inferred from the corresponding paper’s description. }
\label{tab:yolo-fpn-comparison}
\end{table}
\begin{table}[t]
\small
\centering
\setlength{\tabcolsep}{1pt}
\begin{tabular}{lcccccccc}
\toprule
\textbf{{Method}} &\textbf{F. (G)} & \textbf{\#P. (M)} & \textbf{$\text{AP}$} &\textbf{$\text{AP}_{50}$}  &\textbf{$\text{AP}_{75}$}& \textbf{Laten.} \\
\midrule
YOLOv8-N& 8.7  & 3.2  & 37.4 &52.6 &40.5 &1.77 \\
YOLOv10-N &6.7  &\textbf{2.3}  & 38.5 &53.8 &41.7 &1.84 \\
EfficientDet-D1& \textbf{6.1}  & 6.6 & 39.1 & - & - & 1.35 \\
YOLOv11-N & 6.5  & 2.6 &39.4 &55.3 &42.8 & 1.5\\
\rowcolor{Gray}
\textbf{YOLOv11-P$^2$HCT-N}& 10.1  & 2.8 & 40.3 & 56.0 & 43.3 & \textbf{1.24} \\
YOLOv12-N& 6.5 & 2.6 &40.6 &56.7 &43.8 &1.64 \\
\rowcolor{Gray}
\textbf{YOLOv12-P$^2$HCT-N}& 15.8  & 4.3 &\textbf{42.6} &\textbf{59.8} &\textbf{46.1} &2.3 \\
\midrule
YOLOv8-S &28.6  &11.2 &45.0  &61.8 &48.7 &2.33 \\ 
RT-DETR-R18&60.0 &20.0  &46.5 &63.8 &-- &4.58 \\
YOLOv10-S&21.6   &\textbf{7.2} &46.3 &63.0 &50.4 &2.49 \\
YOLOv11-S &21.5 &9.4 &46.9 &63.9 &50.6 &\textbf{2.5}\\
\rowcolor{Gray}
\textbf{YOLOv11-P$^2$HCT-S}& \textbf{21.0}  & 9.3 & \textbf{47.4} & \textbf{64.5} & \textbf{51.3} &\textbf{2.5} \\
\midrule
YOLOv8-M&78.9  &25.9  &50.3 &67.2 &54.7 &5.09\\
RT-DETRv2-R34&100.0 &36.0  &49.9  &67.5 &-- &6.32\\
YOLOv10-M &\textbf{59.1} &\textbf{15.4}  &51.1  &68.1 &55.8 &4.74\\
YOLOv11-M &68.0 &20.1 &51.5 &68.5 &55.7 &4.7\\
\rowcolor{Gray}
\textbf{YOLOv11-P$^2$HCT-M} & 65.9  & 19.5 & \textbf{52.1} & \textbf{69.5} & \textbf{56.7} & \textbf{4.3} \\
\midrule
\end{tabular}
\caption{Comparison with popular state-of-the-art real-time object detectors on MS COCO 2017 validation set. All results are obtained using $640\times640$ inputs. The absence of YOLOv12-P$^2$HCT-S/M data stems from the inability of models trained on these variants to converge, a convergence issue that has also been reported in the original YOLOv12 framework, especially in the classification task. All attention kernel is implemented by FlashAttention. $\text{AP}$s are in the unit of \%, and Latency (Laten.) is in (ms).}
\label{tab:yolo-coco-comparison}
\end{table}
\noindent\textbf{Classification Task:} For classification, we assess the P$^2$HCT-CLS architecture using ResNet-18, ResNet-50, and ResNet-101 as backbones. We also explore lightweight configurations with YOLOv11-cls variants (Nano and Small), replacing the P4 and P5 layer feature fusion structures with P$^2$HCT-CLS. Training is performed on the ImageNet dataset for 200 epochs using SGD with a momentum of 0.9, a learning rate of 0.1, and a batch size of 256. Performance is measured using top-1 and top-5 accuracy.

More parameter settings can be found in supplementary.

\subsection{Detection Performance of P$^2$HCT}
\label{subsec:detection}

For ResNet-based detectors, as shown in Table \ref{tab:yolo-fpn-comparison}, P$^2$HCT-DET consistently outperforms attention-based FPN methods across all scales. At the low-FLOPs scale, the ResNet-18 + P$^2$HCT-N combination achieves an mAP of 46.3\%, markedly exceeding heavier baselines like R-50-AttnFPN (38.4\%) and R-50-A\textsuperscript{2}-FPN-Lite (39.8\%). The superior fine-grained localization is evident with $\text{AP}_{75}$ reaching 49.6\%. At higher FLOPs, ResNet-101 + P$^2$HCT-S achieves an mAP of 50.9\%, significantly exceeding R-101-AttnFPN (40.2\%) and R-101-A\textsuperscript{2}-FPN (42.8\%). The left part of Figure \ref{fig:yolo-tradeoffs} gives a visual comparison of the trade-offs.

For real-time detection, P$^2$HCT-enhanced YOLOv11 models are compared against baselines on the COCO 2017 validation set (Table \ref{tab:yolo-coco-comparison}). YOLOv11-P$^2$HCT-N achieves an mAP of 40.3\%, surpassing the baseline YOLOv11-N (39.4\%) while maintaining a fast latency of 1.24 ms. This performance gain is consistent, with the mid-scale YOLOv11-P$^2$HCT-M reaching 52.1\% mAP, exceeding YOLOv11-M (51.5\%) with comparable or lower FLOPs. Figure \ref{fig:yolo-tradeoffs} (right) further illustrates these accuracy-efficiency trade-offs.

\begin{figure*}[t]
  \centering
  \begin{subfigure}[b]{0.49\textwidth}
    \centering
    \includegraphics[width=0.98\textwidth]{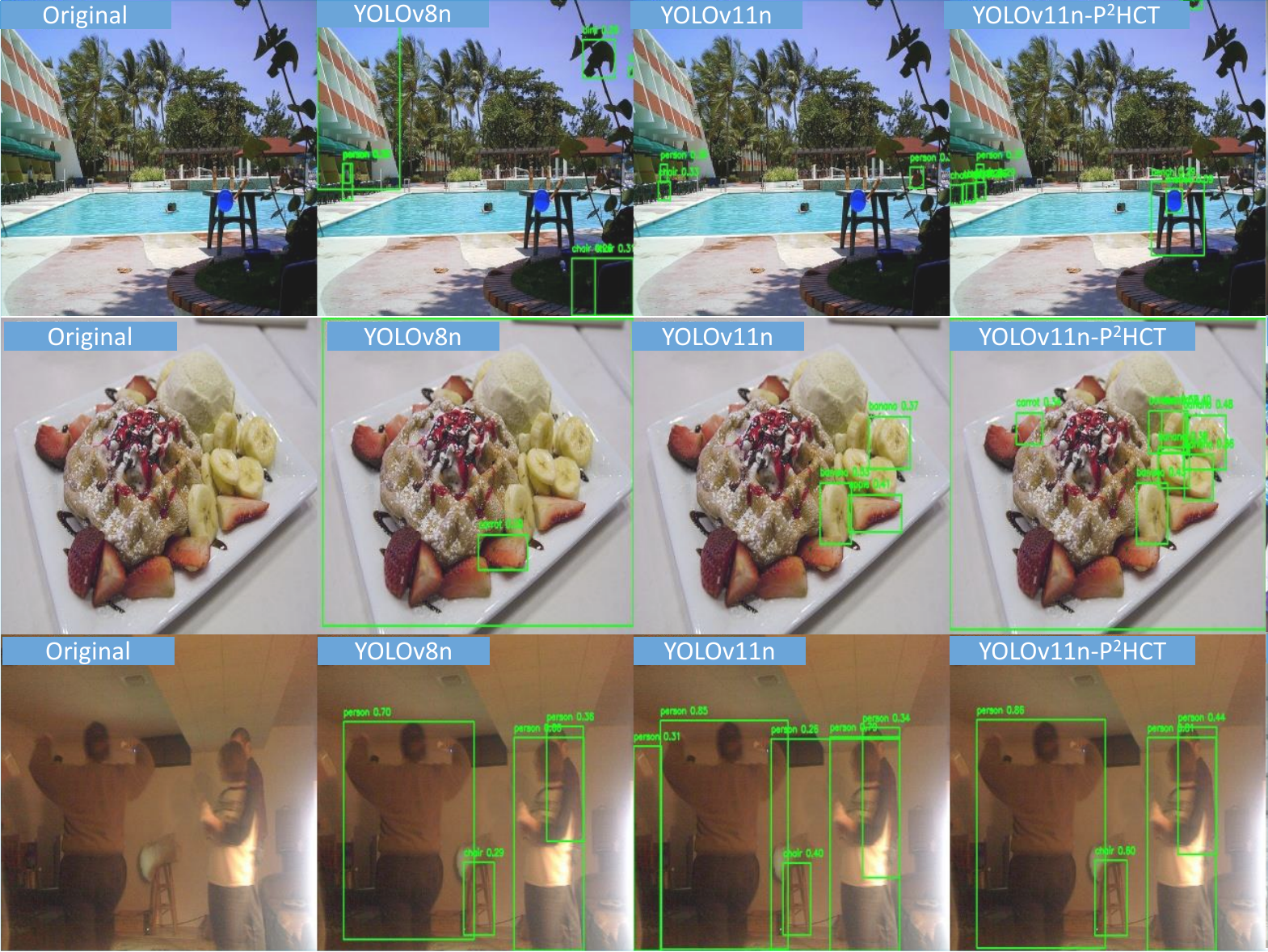}
    \label{fig:visuallization-res}
  \end{subfigure}
  \begin{subfigure}[b]{0.49\textwidth}
    \centering
    \includegraphics[width=0.98\textwidth]{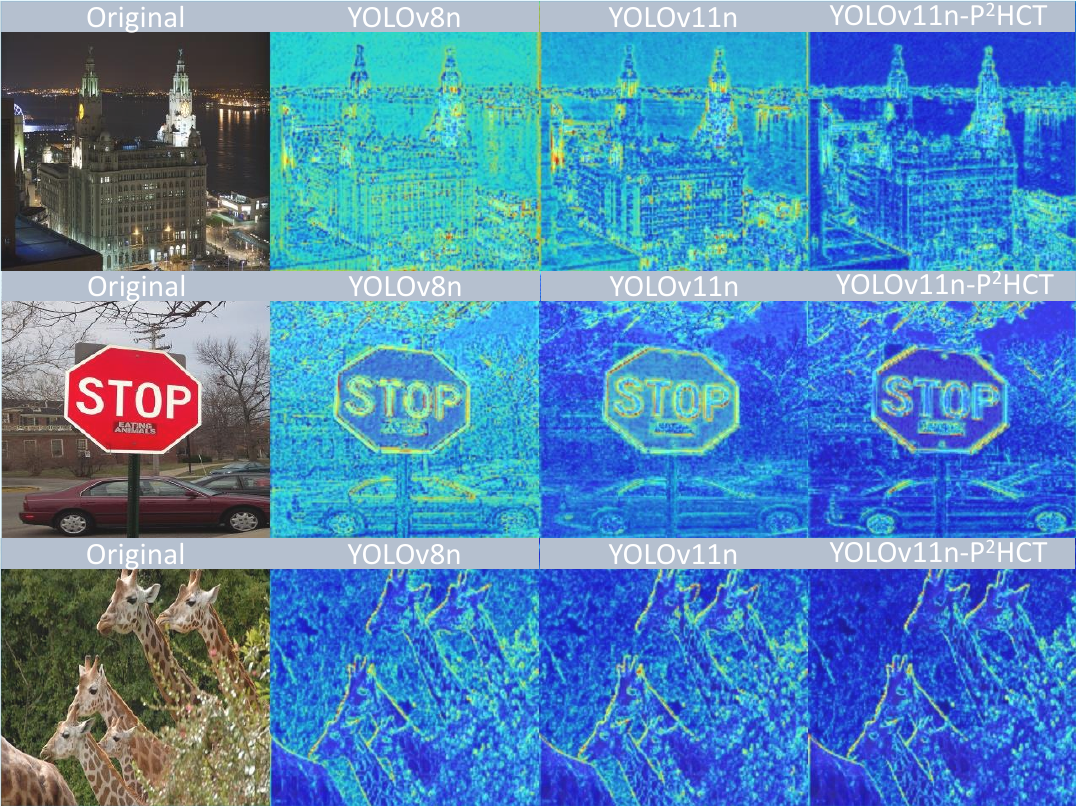}
    \label{fig:heatmap-res}
  \end{subfigure}
  \caption{Object detection results(left) and feature heatmaps(right) of YOLOv8-N, YOLOv11-N, and YOLOv11-P$^2$HCT-N Models on COCO val2017.~Compared to SOTAs, our P$^2$HCT can detect more correct instances with higher scores than other SOTAs and show a purer and cleaner features. [Zoom in for a better view]}
  \label{fig:res}
\end{figure*}

Figure~\ref{fig:res} compares feature heatmaps from YOLOv8-N, YOLOv11-N, and YOLOv11-P$^2$HCT-N. Each row shows an input image followed by the three models’ heatmaps (blue→yellow denotes activation strength). Across varied scenes, P$^2$HCT-enhanced heatmaps appear cleaner and more focused: global coarse attention suppresses irrelevant regions, while fine attention pinpoints precise activations under that guidance.

\subsection{Classification Performance of P$^2$HCT}
We also validate P$^2$HCT as a plug-in for classification (P$^2$HCT-CLS) on ImageNet using ResNet-18/50/101 and lightweight YOLOv11-cls backbones (Table~\ref{tab:comparison-CLS}). Across all architectures, integrating P$^2$HCT yields consistent accuracy gains over the corresponding baselines—for example, ResNet-18 improves by around 6\%, ResNet-50 and ResNet-101 by 1–2\%, and YOLOv11-cls variants by approximately 1\% in top-1 accuracy. These results demonstrate that P$^2$HCT can enhance both heavy and lightweight backbones with minimal overhead, offering a favorable accuracy–efficiency trade-off without requiring a specialized backbone design.

\begin{table}[t]
\centering
\setlength{\tabcolsep}{1pt}
\small
\begin{tabular}{lccccccc}
\toprule
\textbf{{Method}} & \textbf{F. (G)} & \textbf{\#P. (M)} & \textbf{Top1-Acc} &\textbf{Top5-Acc}  \\
\midrule
YOLOv8-cls-N& 4.9  & 2.7  & 69.0  & 88.3  \\
YOLOv11-cls-N& 5.3  &1.6   &70.0 &89.4  \\
\rowcolor{Gray}
\textbf{YOLOv11-cls-P$^2$HCT-N}& 4.9  & 3.5 & 70.9  & 89.6  \\
EfficientNet-B0 & 3.9  & 5.3 & 77.1  & 93.3  \\
\midrule
R-18 & 42  & 17.9 & 68.5  & -  \\
\rowcolor{Gray}
\textbf{R-18-P$^2$HCT-N}& 43  & 19.6 & 75.0  & 92.5  \\
PVTv2-b1 & 21  & 13.1 & 78.7  & - \\
YOLOv8-cls-S& 17  & 6.4  & 73.8  & 91.7  \\
YOLOv11-cls-S &16  &5.5   &75.4 &92.7  \\
\rowcolor{Gray}
\textbf{YOLOv11-cls-P$^2$HCT-S}& 19 & 11.5 & 76.2  & 93.0 \\
\midrule
R-50& 71  & 27.4 & 76.2  & -  \\
\rowcolor{Gray}
\textbf{R-50-P$^2$HCT-N}~ & 72  & 29.2 & 77.9 & 94.0  \\
Swin-T& 45  & 28.3 & 81.2  & -  \\
PVTv2-b2 & 40  & 25.4 & 82.0  & -  \\
\midrule
R-101& 132  & 46.4 & 77.4  & -  \\
\rowcolor{Gray}
\textbf{R-101-P$^2$HCT-S}~ & 140  & 54.4 & 78.4 & 94.1  \\
ViT-B & 176  & 86.6 & 77.9  & -  \\
Swin-S& 87  & 49.6 & 83.1 & -  \\
Swin-B & 154  & 87.8 & 83.4  & -  \\
\midrule
\end{tabular}
\caption{Comparison with popular SOTA image classifiers on the ImageNet dataset. All results are obtained using $224\times224$ inputs. Top-1 Accuracy and Top-5 Accuracy are in (\%).}
\label{tab:comparison-CLS}
\end{table}

\subsection{Ablation Study}
To assess the contributions of key components in the P$^2$HCT, we conduct ablation experiments on the COCO 2017 dataset, focusing on the detection task using the YOLOv11-P$^2$HCT-N model. The experiments evaluate the effects of top-\(k\) selection, P2HCA stacked design, self-gating mechanism, linear attention variants, parameter sharing, and attention kernel types. Results are summarized in Table \ref{tab:ablation_results}, with bold entries indicating the best-performing configurations.

\textbf{Top-\(k\) Selection:} An intriguing finding is that replacing top-\(k\) with Gumbel-Softmax soft top-\(k\) \cite{rao2021dynamicvit} and enabling fine attention during training resulted in significant training difficulties and a precision drop exceeding 5\%. Table \ref{tab:diagnosis_attention_implementation} presents results with fine attention enabled solely during the inference phase, it shows that a top-\(k\) value of 8 achieves the highest AP\(_{50:95}\) of 40.3\% with a latency of 1.24 ms. Increasing top-\(k\) to 16 and 32 reduces AP\(_{50:95}\) to 39.5\% and 38.9\%, respectively, while significantly increasing latency to 1.50 ms and 1.94 ms. This indicates that selecting a moderate number of top-\(k\) tokens (e.g., 8) optimizes both accuracy and efficiency, whereas larger values introduce computational overhead without performance gains.

\textbf{P2HCA Stacked Design:} The P2HCA stacked structure operates such that the query ($Q$) of a P2HCA module receives input from the previous P2HCA layer, while the key ($K$) and value ($V$) remain consistent across all P2HCA modules. Finally, the outputs of all P2HCA modules are concatenated for subsequent processing. Table \ref{tab:diagnosis_stack_design} reveals that a single P2HCA stage yields the best AP\(_{50:95}\) of 40.3\% with the lowest latency of 1.24 ms. Stacking additional stages (2, 4, 6) decreases AP\(_{50:95}\) to 39.8\% and 39.6\% while dramatically increasing latency to 2.17 ms, 3.65 ms, and 6.17 ms, respectively. This suggests that a single P2HCA stage is sufficient for effective feature fusion, while additional stages add unnecessary complexity without accuracy gains.

\textbf{Self-Gating Mechanism:} Following the inspiration of NSA \cite{yuan2025native}, the self-gating mechanism is designed to fuse coarse and fine attention outputs using a gating tensor. Given coarse out $O_{\mathrm{coarse}}$ and refined out $O_{\mathrm{refined}}$ with the same shape, we concatenate them and apply a 1D convolution followed by a sigmoid layer $\sigma(\cdot)$:
\begin{equation}
g = \sigma(\text{Conv1d}(\text{Concat}(O_{\mathrm{coarse}}, O_{\mathrm{fine}}))).
\end{equation}
Thus, the final output is:
\begin{equation}
\begin{aligned}
  O = \mathrm{Conv1x1}(&g \cdot O_{\mathrm{fine}} + (1 - g) \cdot O_{\mathrm{coarse}} \\
  &+ \mathrm{Upsample}(\mathrm{DConv7x7}(V))).  
\end{aligned}
\end{equation}
Table \ref{tab:diagnosis_self_gating} demonstrates that disabling the self-gating mechanism improves AP\(_{val}\) for both Nano (N) and Small (S) models, achieving 40.3\% and 47.4\%, respectively, compared to 40.0\% and 46.2\% with gating enabled. This indicates that the self-gating mechanism may introduce unnecessary complexity, slightly degrading performance.


\newcommand{\subtablewidth}{0.2\textwidth}
\begin{table}[!t]
\small
\centering
\setlength{\tabcolsep}{0.06cm}
\begin{subtable}[t]{\subtablewidth}
\centering
\begin{tabular}[t]{ccc}
\toprule
\textbf{Top-k} & \textbf{$\text{AP}^{val}_{50:95}$ } & \textbf{Laten.} \\
\midrule
0 & 40.2  & 1.21 \\
\rowcolor{Gray}
8 & 40.3  & 1.24 \\
16 &  39.5 & 1.50 \\
32 &  38.9 & 1.94 \\
\bottomrule
\end{tabular}
\caption{Index of Top-k}
\label{tab:diagnosis_attention_implementation}
\end{subtable}
~~~~~~
\begin{subtable}[t]{\subtablewidth}
\centering
\begin{tabular}[t]{ccc}
\toprule
\textbf{Num.} & \textbf{$\text{AP}^{val}_{50:95}$ } & \textbf{Laten.} \\
\midrule
\rowcolor{Gray}
1 & 40.3  & 1.24  \\
2 & 39.8  & 2.17 \\
4   & 39.6  & 3.65 \\
6   & 39.6  & 6.17 \\
\bottomrule
\end{tabular}
\caption{P2HCA Stacked Design}
\label{tab:diagnosis_stack_design}
\end{subtable}
\\
\begin{subtable}[t]{\subtablewidth}
\centering
\begin{tabular}[t]{ccc}
\toprule
\textbf{SG} & \textbf{$\text{AP}^{val}$ (N) } & \textbf{$\text{AP}^{val}$ (S)}\\
\midrule
w SG & 40.0  & 46.2 \\
\rowcolor{Gray}
w/o SG   & 40.3  & 47.4 \\
\bottomrule
\end{tabular}
\caption{Self-Gating mechanism.}
\label{tab:diagnosis_self_gating}
\end{subtable}
~~~~~~~
\begin{subtable}[t]{\subtablewidth}
\centering
\begin{tabular}[t]{ccc}
\toprule
\textbf{Method} & \textbf{$\text{AP}^{val}_{50:95}$ } & \textbf{Laten.}\\
\midrule
\rowcolor{Gray}
w & 40.3 & 1.24 \\
w/o & 40.3 & 1.24 \\
\bottomrule
\end{tabular}
\caption{Parameter Sharing}
\label{tab:diagnosis_parameter_share}
\end{subtable}
\caption{\textbf{Ablation studies.} We only show the factor(s) to be diagnosed in each subtable to save space. The default parameters are trained for $600$ epochs from scratch, using YOLOv11-P$^2$HCT-N model on MS COCO 2017 dataset. $\text{AP}$s are in the unit of (\%), and Latency (Laten.) is in (ms).}
\label{tab:ablation_results}
\end{table}

\textbf{Parameter Sharing:} In the design of P2HCA, both coarse attention and fine attention share the same set of convolutions to compute the key and value. Here, we compare this with the scenario where they are computed separately. Table \ref{tab:diagnosis_parameter_share} indicates that parameter sharing between coarse and fine attention stages does not affect AP\(_{50:95}\) (40.3\% for both shared and unshared settings) or latency (1.24 ms). However, parameter sharing reduces the model’s parameter count and training complexity, making it a favorable design choice without sacrificing performance.


\section{Conclusion}
\label{sec:conclusion}
We introduced the \textbf{PnP Hierarchical C2F Transformer (P$^2$HCT)}, a novel and versatile module that resolves the high computational cost of feature fusion by employing a coarse-to-fine cross-attention strategy and an efficient parameter-sharing scheme. Demonstrated as a seamless, plug-and-play component, P$^2$HCT consistently enhanced performance across various backbone networks, delivering significant accuracy gains on the COCO and ImageNet datasets in object detection (P$^2$HCT-DET) and image classification (P$^2$HCT-CLS) while maintaining low computational overhead. This work validates that superior hierarchical feature fusion can be achieved through highly efficient, sparse, and content-adaptive attention mechanisms, offering a practical solution for resource-constrained vision tasks.

\bibliographystyle{IEEEbib}
\bibliography{icme2026references}

\appendix

\section*{Coarse to fine and approximate theoretical support for the rationality of shared parameters}
\subsection*{Problem Setup and Notations}
To justify the Coarse to Fine setup and parameter sharing, we examine the following simplified model. We replace the standard attention mechanism with Linear Attention. For input features, given each sample $i=1,\dots,N$, let $\mathbf U_i \in \mathbb R^d$. For Linear Attention, the shared parameters are:
\[
      W_Q,\;W_K,\;W_V \iid \mathcal N\bigl(\mathbf 0,\;\boldsymbol{\Sigma}\bigr).
\]

The output of the coarse branch (Coarse Attention) is:
\[
    O_{c,i}= O_c(\mathbf U_i)
      \;=\;(W_Q\mathbf U_i)\;(W_K\mathbf U_i)^\trsp\;(W_V\mathbf U_i)
      \;\in\;\mathbb R^d.
\]

Define the internal importance score for the coarse branch:
\[
  r_i \;=\;\frac1d\sum_{j=1}^d\bigl[\,Q\,K^\trsp\bigr]_{ij} = \frac1d\sum_{j=1}^d\bigl[(W_Q\mathbf U_i)\;(W_K\mathbf U_i)^\trsp\bigr]_{ij},
\]
Denote its order statistics as $r_{(1)}\le\cdots\le r_{(d)}$, and select the index set of the top-$k$ rows:
\[
  \mathcal S \;=\;\bigl\{\,i : r_i \ge r_{(d-k+1)}\bigr\},
  \quad
  S = \mathrm{diag}\bigl(\mathbf 1_{i\in\mathcal S}\bigr).
\]
The fine branch (Fine Attention) output is then defined as:
\[
\begin{aligned}
    O'_{f,i}
  &= Q_i\,(S\,K_f^\trsp\,V_f)_i\\
  &=\;(W_Q\mathbf U_i)\;\Bigl(\sum_{j\in\mathcal S}(W_K\mathbf U_i)_j\,(W_V\mathbf U_i)_j\Bigr).
\end{aligned}
\]
Considering the output $O_{c,i} + O_{f,i}$ with MSE loss, and letting $\mathbf Y_i$ be the expected (label) output for sample $i$, we define:
\[
    \begin{aligned}
      L_c &= \frac1N\sum_{i=1}^N\|\mathbf Y_i - O_{c,i}\|^2
            = \frac1N\sum_i\|\mathbf R_i\|^2,\\
      L_{c+f} &= \frac1N\sum_{i=1}^N\|\mathbf Y_i - O_{c,i} - O'_{f,i}\|^2
              = \frac1N\sum_i\|\mathbf R_i - O'_{f,i}\|^2.
    \end{aligned}
\]
The core problem is to examine the difference term for a structure pre-trained using $L_c$:
\[
      \Delta L \;=\; L_c - L_{c+f}
      \;=\;\frac2N\sum_i \mathbf R_i^\trsp O'_{f,i}
        \;-\;\frac1N\sum_i \|O'_{f,i}\|^2.
\]
and the probability that $\expe{}{\Delta L} \geq 0$. 

\subsection*{Well-Trained Conditions and Some Lemmas}
\begin{definition}[Well-Trained Condition]\label{def:welltrain}
    Following previous notations. Consider an orthogonal decomposition of operator $O_c$, i.e., there exists an orthogonal projection operator $O_r$ such that:
    $$
        O_c\cdot O_r = \boldsymbol{0},\,\, O_c+O_r = \mathbf I.
    $$
    The image spaces of $O_c$ and $O_f$ are denoted as $V_c = \mathrm{Im}(O_c)$ and $V_r = \mathrm{Im}(O_r)$, respectively.

    For each sample's expected (label) output $\mathbf Y_i$, it can be decomposed as $\mathbf Y_i = \mathbf Y_{c,i} + \mathbf Y_{r,i}$, where $\mathbf Y_{c,i}\in V_c$ and $\mathbf Y_{r,i}\in V_r$.
    
    If the network is well-trained under the $L_c$ output, then:
    $$
        O_{c,i} = O_c(\mathbf U_i) = \mathbf Y_{c,i}.
    $$
    Simultaneously, $\boldsymbol{\Sigma}$ can be decomposed as:
    $$
        \boldsymbol{\Sigma} = \alpha \boldsymbol{\Sigma}_c + \beta \boldsymbol{\Sigma}_r.
    $$
\end{definition}

\begin{lemma}\label{lma:u2r}
Following previous notations and assuming well-trained conditions, Definition \ref{def:welltrain} directly gives:
$$
    \mathbf R_i = \mathbf Y_i-\mathbf O_{c,i} = \mathbf Y_{c,i} + \mathbf Y_{r,i} - \mathbf O_{c,i} = \mathbf Y_{r,i} = O_r(\mathbf U_i)\in V_r.
$$
$\forall i$, consider the attention difference:
\[
\Delta_i \;=\;\langle \mathbf R_i,\;O_c(\mathbf U_i)\rangle \;-\;\langle \mathbf R_i,\;O_c(\mathbf R_i)\rangle.
\]
Then there exist constants \(C,c>0\) such that for any \(t>0\):
\[
\Pr\Bigl(
\bigl|\Delta_i\bigr|
\;\ge\;
C\,|\boldsymbol{\Sigma}|^3(\sqrt{m}+t)^3\Bigr) \leq 6e^{-c\,t^2}.
\]
In particular, when \(t=\sqrt m\), the bound becomes \(6e^{-c m}\).
\end{lemma}

\begin{proof}
Substitute \(\mathbf U_i=O_c\mathbf U_i+\mathbf R_i\) into
\(O_c(\mathbf U_i)-O_c(\mathbf R_i)\), retaining only cross terms containing \(O_c\mathbf U_i\):
\[
\begin{aligned}
    O_c(\mathbf U_i)-O_c(\mathbf R_i)
&=\sum_{\ell=1}^3
W_Q O_c(\mathbf U_i)\,(W_K\mathbf R_i)^\trsp\,(W_V\mathbf R_i)\\
& +\;\text{(other minor terms)}.
\end{aligned}
\]
Thus:
\[
\begin{aligned}
    |\Delta_i|
&=\bigl|\langle \mathbf R_i,\;O_c(\mathbf U_i)-O_c(\mathbf R_i)\rangle\bigr|\\
& \leq \|\mathbf R_i\|\sum_{\ell=1}^3
\bigl\|W_Q O_c(\mathbf U_i)\bigr\|\,
\bigl\|W_K\mathbf R_i\bigr\|\,
\bigl\|W_V\mathbf R_i\bigr\|.
\end{aligned}
\]
For any fixed vector \(\mathbf Z\), Gaussian matrix mappings satisfy the sub-Gaussian norm concentration inequality:  
There exists \(c'>0\) such that for all \(s>0\):
\[
\Pr\Bigl(\|W_i \mathbf Z\|\ge|\boldsymbol{\Sigma}|\|\mathbf Z\|(\sqrt m + s)\Bigr)
\;\le\;
e^{-c's^2},
 i\in\{Q,K,V\}.
\]
Applying a union bound over the three factors and the pure coarse term (totaling at most 6 terms), we obtain:
\[
\begin{aligned}
    \Pr\Bigl(
\exists\,\ell &:\;W_Q O_c(\mathbf U_i)\,(W_K\mathbf R_i)^\trsp\,(W_V\mathbf R_i)\\
&\ge |\boldsymbol{\Sigma}|^3\|O_c(\mathbf U_i)\|\|\mathbf R_i\|^2(\sqrt m + t)^3
\Bigr) \leq 6e^{-c\,t^2}.
\end{aligned}
\]
Since $\|O_c(\mathbf U_i)\|\|\mathbf R_i\|^2$ is clearly bounded, there exists $C>0$ satisfying the lemma.
\end{proof}

\begin{lemma}\label{lma:ofi}
Following previous notations, given a vector \(u\in\mathbb{R}^d\). Define
\[
a = u^\trsp Q\,u,\qquad w_i = (V\,u)_i.
\]
Let
\[
O'_{f,i} = u^\trsp Q\,u\;\Bigl(\sum_{i\in S} r_i\,w_i\Bigr).
\]
Denote
\[
\begin{aligned}
    \nu^2& = \mathrm{Var}(r_i) \;=\;\frac{\mathrm{Tr}(\boldsymbol{\Sigma}^2)}{d},\\
\mu &= \mathrm{Cov}(a,\,r_i) \;=\;\frac{u^\trsp \boldsymbol{\Sigma}\,u\;\mathrm{Tr}(\boldsymbol{\Sigma})}{d},
\end{aligned}
\]
and let $\Phi^{-1}$ be the standard normal quantile function. Then
\[
\begin{aligned}
    \mathbb{E}[O'_{f,i}]
&= \frac{\mu\,\nu}{k}
\Bigl(\sum_{m=d-k+1}^d \Phi^{-1}\!\bigl(\tfrac{m}{d+1}\bigr)\Bigr)
\;\times\;(u^\trsp V\,u)\\
\;&>\mathbb{E}[O_{f,i}]=0.
\end{aligned}
\]
\end{lemma}

\begin{proof}
Since \((a,r_i)\) are zero-mean joint Gaussians with
\[
\mathrm{Cov}(a,r_i)=\mu,\quad\mathrm{Var}(r_i)=\nu^2,
\]
for any \(x\in\mathbb{R}\) we have
\[
\mathbb{E}[\,a\,r_i\mid r_i=x]
=\frac{\mu}{\nu^2}\,x.
\]

Thus:
\[
\begin{aligned}
    \mathbb{E}[O'_{f,i}]
&=\sum_{i=1}^d w_i\;\mathbb{E}\bigl[a\,r_i\,\mathbf1_{i\in S}\bigr]\\
&=\sum_{i=1}^d w_i\;\frac{\mu}{\nu^2}\,\mathbb{E}\bigl[r_i\,\mathbf1_{r_i\ge r_{(n-k+1)}}\bigr].
\end{aligned}
\]

For \(r_i\sim\mathcal{N}(0,\nu^2)\), the expectation of the $m$-th largest value is
\(\mathbb{E}[r_{(m)}]=\nu\,\Phi^{-1}(m/(n+1))\).  
Hence
\[
\begin{aligned}
    \mathbb{E}\bigl[r_i\,\mathbf1_{r_i\ge r_{(d-k+1)}}\bigr]
&=\frac{1}{k}\sum_{m=d-k+1}^d\mathbb{E}[r_{(m)}]\\
&=\frac{\nu}{k}\sum_{m=d-k+1}^d\Phi^{-1}\!\Bigl(\tfrac{m}{d+1}\Bigr).
\end{aligned}
\]

Therefore:
\[
\begin{aligned}
    \mathbb{E}[O'_{f,i}]
&=\frac{\mu}{\nu^2}\sum_{i=1}^d w_i\;
\frac{\nu}{k}\sum_{m=d-k+1}^d\Phi^{-1}\!\Bigl(\tfrac{m}{d+1}\Bigr)\\
&=\frac{\mu\,\nu}{k}\Bigl(\sum_{m=d-k+1}^d \Phi^{-1}\!\tfrac{m}{d+1}\Bigr)\,(u^\trsp V\,u).
\end{aligned}
\]
Since \(\sum\Phi^{-1}(m/(d+1))>0\) and \(\mu,\nu>0\), we have \(\mathbb{E}[O'_{f,i}]>0\).
\end{proof}

\begin{lemma}\label{lma:salpha}
Following previous notations, let \(\boldsymbol{\Sigma}_c\) and \(\boldsymbol{\Sigma}_r\) be two positive semidefinite matrices satisfying
\[
\boldsymbol{\Sigma}_c\,\boldsymbol{\Sigma}_r = 0,
\]
and set 
\(\boldsymbol{\Sigma} = \alpha\,\boldsymbol{\Sigma}_r + \beta\,\boldsymbol{\Sigma}_c\)
for scalars \(\alpha,\beta\ge0\).  Let \(q\sim\mathcal N(\mathbf0,\boldsymbol{\Sigma})\) be a Gaussian vector, and let \(u\) lie in the image of \(\boldsymbol{\Sigma}_r\), i.e.\ \(\boldsymbol{\Sigma}_c\,u=0\).  Then
\[
\mathbb{E}\bigl[(q^\trsp u)^2\bigr] \;=\;\alpha\,u^\trsp \boldsymbol{\Sigma}_r\,u.
\]
\end{lemma}

\begin{proof}
Write the covariance decomposition
\(\boldsymbol{\Sigma}=\alpha\,\boldsymbol{\Sigma}_r+\beta\,\boldsymbol{\Sigma}_c\).
Since \(q\sim N(0,\boldsymbol{\Sigma})\), for any fixed \(u\in\mathbb R^d\),
\[
\begin{aligned}
    \mathbb{E}\bigl[(q^\trsp u)^2\bigr]
&= u^\trsp \boldsymbol{\Sigma}\,u
= u^\trsp\bigl(\alpha\,\boldsymbol{\Sigma}_r + \beta\,\boldsymbol{\Sigma}_c\bigr)u\\
&= \alpha\,u^\trsp\boldsymbol{\Sigma}_r\,u
\;+\;\beta\,u^\trsp\boldsymbol{\Sigma}_c\,u.
\end{aligned}
\]
Noticed that accroding to definition \ref{def:welltrain}, \(u\) lies in \(\mathrm{Im}(\boldsymbol{\Sigma}_r)\) and is orthogonal to \(\mathrm{Im}(\boldsymbol{\Sigma}_c)\), so
\(\boldsymbol{\Sigma}_c\,u=0\).  Hence
\(\,u^\trsp\boldsymbol{\Sigma}_c\,u=0\), and the formula reduces to
\[
\mathbb{E}\bigl[(q^\trsp u)^2\bigr]
= \alpha\,u^\trsp\boldsymbol{\Sigma}_r\,u.
\]
This completes the proof.
\end{proof}

\subsection*{Conclusion}
\begin{theorem}
Following previous notations, if the decomposition coefficient satisfies $\alpha < 2/|\boldsymbol{\Sigma}_r|$, then there exist $C>0$ and $\delta \in (0,1)$ such that:
\[
    \prob{\expe{}{\Delta L} \geq 0} \geq 1- e^{-C d_r \delta^2},
\]
where $d_r = \mathrm{Rank(\boldsymbol{\Sigma}_r)}$.
\end{theorem}
\begin{proof}
\begin{equation*}
        \begin{aligned}
            \Delta L &= L_c - L_{c+f} = \frac2N \sum_i \mathbf R_i^\trsp\mathbf O'_{f,i} - \frac1N \sum_i \|\mathbf O'_{f,i}\|^2\\
            & \overset{\text{Lemma }\ref{lma:ofi}}{\geq }\frac2N \sum_i \mathbf R_i^\trsp\mathbf O_{f,i} - \frac1N \sum_i \|\mathbf O'_{f,i}\|^2.
        \end{aligned}
    \end{equation*}

Since $\mathbf R_i^\trsp\mathbf O_{f,i}$ is sub-Gaussian, by concentration inequality, there exists $c_1\geq 0$ such that $\forall \delta_1 \in (0,1)$:
\begin{equation*}
    \Pr\Bigl(|\mathbf R_i^\trsp\mathbf O_{f,i} - \expe{}{\mathbf R_i^\trsp\mathbf O_{f,i}}| \geq \delta_1 \expe{}{\mathbf R_i^\trsp\mathbf O_{f,i}}\Bigr) \leq 2e^{-c_1 d_r \delta_1^2}.
\end{equation*}

We estimate:
\begin{equation*}
    \begin{aligned}
        &\Pr\Bigl(\expe{}{\mathbf R_i^\trsp\mathbf O_{f,i}} - \expe{}{\mathbf R_i^\trsp(W_Q \mathbf R_i)\cdot(W_K \mathbf R_i)^\trsp\cdot(W_V \mathbf R_i)} \\
        &\geq c_2|\boldsymbol{\Sigma}|^3(\sqrt{d}+\delta_2)^3\Bigr) \overset{\text{Lemma }\ref{lma:u2r}}{\leq}  6e^{-c_2\delta_2^2}.
    \end{aligned}
\end{equation*}

Note that:
\begin{equation*}
    \begin{aligned}
        &\expe{}{\mathbf R_i^\trsp(W_Q \mathbf R_i)\cdot(W_K \mathbf R_i)^\trsp\cdot(W_V \mathbf R_i)} \\
        &= \expe{}{\mathbf R_i^\trsp(W_Q \mathbf R_i)}\expe{}{(W_K \mathbf R_i)^\trsp\cdot(W_V \mathbf R_i)}\\
        & \overset{\text{lemma}\ref{lma:salpha}}{=} \alpha \|\mathbf R_i\|^2|\boldsymbol{\Sigma}_r|\cdot \alpha \|\mathbf R_i\|^2|\boldsymbol{\Sigma}_r| = \alpha^2 \|\mathbf R_i\|^4|\boldsymbol{\Sigma}_r|^2.
    \end{aligned}
\end{equation*}

Combining the above expressions:
$$
\begin{aligned}
    \Pr&\Bigl(\mathbf R_i^\trsp\mathbf O_{f,i} \geq (1-\delta_1) \alpha^2|\boldsymbol{\Sigma}_r|^2\|\mathbf R_i\|^4\Bigr) \\
    &\leq 1-8e^{c_1c_2d_r \min\{\delta_1,\delta_2\}^2}.
\end{aligned} 
$$

Similarly, there exists $c_3\geq 0$ such that $\forall \delta_3 \in (0,1)$:
$$
    \Pr\Bigl(\|\mathbf O'_{f,i}\| \leq (1+\delta_3)\alpha^3 |\boldsymbol{\Sigma}_r|^3\|\mathbf R_i\|^4\Bigr) \leq 1-2e^{-c_3 d_r \delta_3^2}.
$$

Therefore:

$$
\begin{aligned}
    \Pr\Bigl(&\expe{}{\Delta L_i} \geq 2(1-\delta_1) \alpha^2|\boldsymbol{\Sigma}_r|^2\|\mathbf R_i\|^4 \\
    &- (1+\delta_3)\alpha^3 |\boldsymbol{\Sigma}_r|^3\|\mathbf R_i\|^4 \geq 0 \Bigr) \\
    &\geq 1- e^{-c_1c_2c_3 d_r \min\{\delta_1,\delta_2,\delta_3\}^2}.
\end{aligned}
$$
When $\alpha \leq \frac{2(1-\delta_1)}{(1+\delta_3)|\boldsymbol{\Sigma}_r|} < 2/|\boldsymbol{\Sigma}_r|$, taking $C = c_1c_2c_3$ and $\delta = \min\{\delta_1,\delta_2,\delta_3\}$ completes the proof.

\end{proof}
\section*{Limitations} 

The Plug-and-Play Hierarchical C2F Transformer demonstrates notable improvements in multi-scale feature fusion but faces several limitations. First, regarding the top-\(k\) selection strategy, one might intuitively expect that enabling fine attention during training would yield better results; however, this is not the case, suggesting underlying mechanisms that warrant further exploration. Although effective in this specific context, bypassing the joint training of the fine-grained attention branch via top-k selection during training is likely not a universally optimal solution. Additionally, the default top-\(k\) value of 8 is largely empirical in nature. Also, P$^2$HCT benefits from optimized libraries like FlashAttention or SageAttention, which require modern GPUs, potentially limiting accessibility. 

\section*{More Details}
All P$^2$HCT-DET and P$^2$HCT-CLS models are trained using the SGD optimizer, with P$^2$HCT-DET-N/S/M models trained for 600 epochs and P$^2$HCT-CLS-ALL models for 200 epochs, as detailed in Table \ref{tab:finetuning_details}. Following previous works, the SGD momentum is set to 0.937 for P$^2$HCT-DET and 0.9 for P$^2$HCT-CLS, with weight decay values of \(5 \times 10^{-4}\) and \(1 \times 10^{-4}\), respectively. The initial learning rate for P$^2$HCT-DET is \(1 \times 10^{-2}\), decaying linearly to \(1 \times 10^{-4}\), while P$^2$HCT-CLS starts at 0.2 and decays to \(2 \times 10^{-3}\). Data augmentations, including Mosaic, Mixup, and copy-paste augmentation, are applied to enhance training for P$^2$HCT-DET, with additional HSV, translation, and scale augmentations for both tasks. All models are trained on \(8 \times\) NVIDIA RTX 4090 GPUs, and we report the standard mean average precision (mAP). 

\begin{table*}[h]
    \centering
    \setlength{\tabcolsep}{0.18cm}
    \caption{\textbf{Hyperparameters of P$^2$HCT-DET and P$^2$HCT-CLS training settings.}}
    \begin{tabular}{l|cc}
    \toprule
    \textbf{Hyperparameters}                   & \textbf{P$^2$HCT-DET-N/S/M} & \textbf{P$^2$HCT-CLS-ALL}\\ 
    \midrule
    \multicolumn{3}{l}{\textbf{\textit{Training Configuration}}} \\
    Epochs                                    & $600$ & $200$ \\
    Optimizer                                 & SGD & SGD \\
    Momentum                                  & $0.937$ & $0.9$ \\
    Batch size                                & $32\times8$ & $32\times8$ \\
    Weight decay                              & $5 \times 10^{-4}$ & $1 \times 10^{-4}$\\
    Warm-up epochs                            & $3$ & $0$ \\
    Warm-up momentum                          & $0.8$ & - \\
    Warm-up bias learning rate                & $0.0$ & - \\
    Initial learning rate                     & $10^{-2}$ & $0.2$ \\
    Final learning rate                       & $10^{-4}$ & $2\times 10^{-3}$\\
    Learning rate schedule                    & Linear decay & Linear decay \\
    Automatic mixed precision                 & True & True \\
    \midrule
    \multicolumn{3}{l}{\textbf{\textit{Loss Parameters}}} \\
    Box loss gain                             & $7.5$ & - \\
    Class loss gain                           & $0.5$ & - \\
    DFL loss gain                             & $1.5$ & - \\
    \midrule
    \multicolumn{3}{l}{\textbf{\textit{Augmentation Parameters}}} \\
    HSV saturation augmentation               & $0.7$ & $0.4$ \\
    HSV value augmentation                    & $0.4$ & $0.4$ \\
    HSV hue augmentation                      & $0.015$ & $0.015$\\
    Translation augmentation                  & $0.1$ & $0.1$  \\
    Scale augmentation                        & $0.5 / 0.9 / 0.9 $ & $0.5$\\
    Mosaic augmentation                       & $1.0$ & -\\
    Mixup augmentation                        & $0.0 / 0.05 / 0.15 $ & $0.0$\\
    Copy-paste augmentation                   & $0.1 / 0.15 / 0.4 $ & $0.0$\\
    Close mosaic epochs                       & $10$ & -\\
    \midrule
    \multicolumn{3}{l}{\textbf{\textit{P$^2$HCT Parameters}}} \\
    $C'$(token dimension)                     & $64,128/128,256/256,512$ & $256/1024/1024$\\
    MLP extension factor                       & $2$ & $2$\\
    \bottomrule
    \end{tabular}
    \label{tab:finetuning_details}
\end{table*}

\section*{Additional Detection Visualizations and Attention Heatmaps}
\begin{figure*}
    \centering
    \includegraphics[width=1\linewidth]{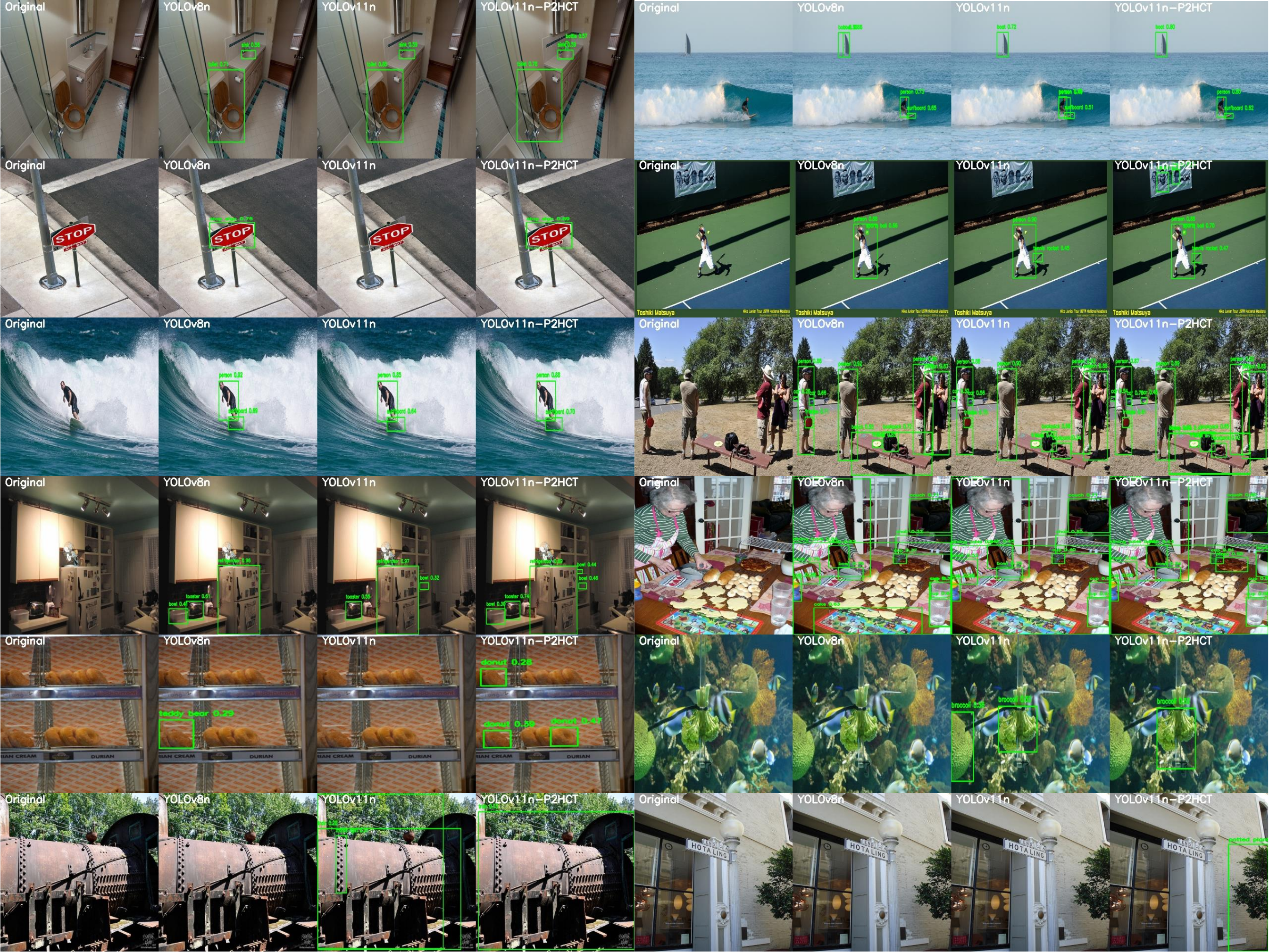}
    \caption{Example detection results of different methods on COCO 2017 validation set.}
    \label{fig:supp-visual}
\end{figure*}

\begin{figure*}
    \centering
    \includegraphics[width=1\linewidth]{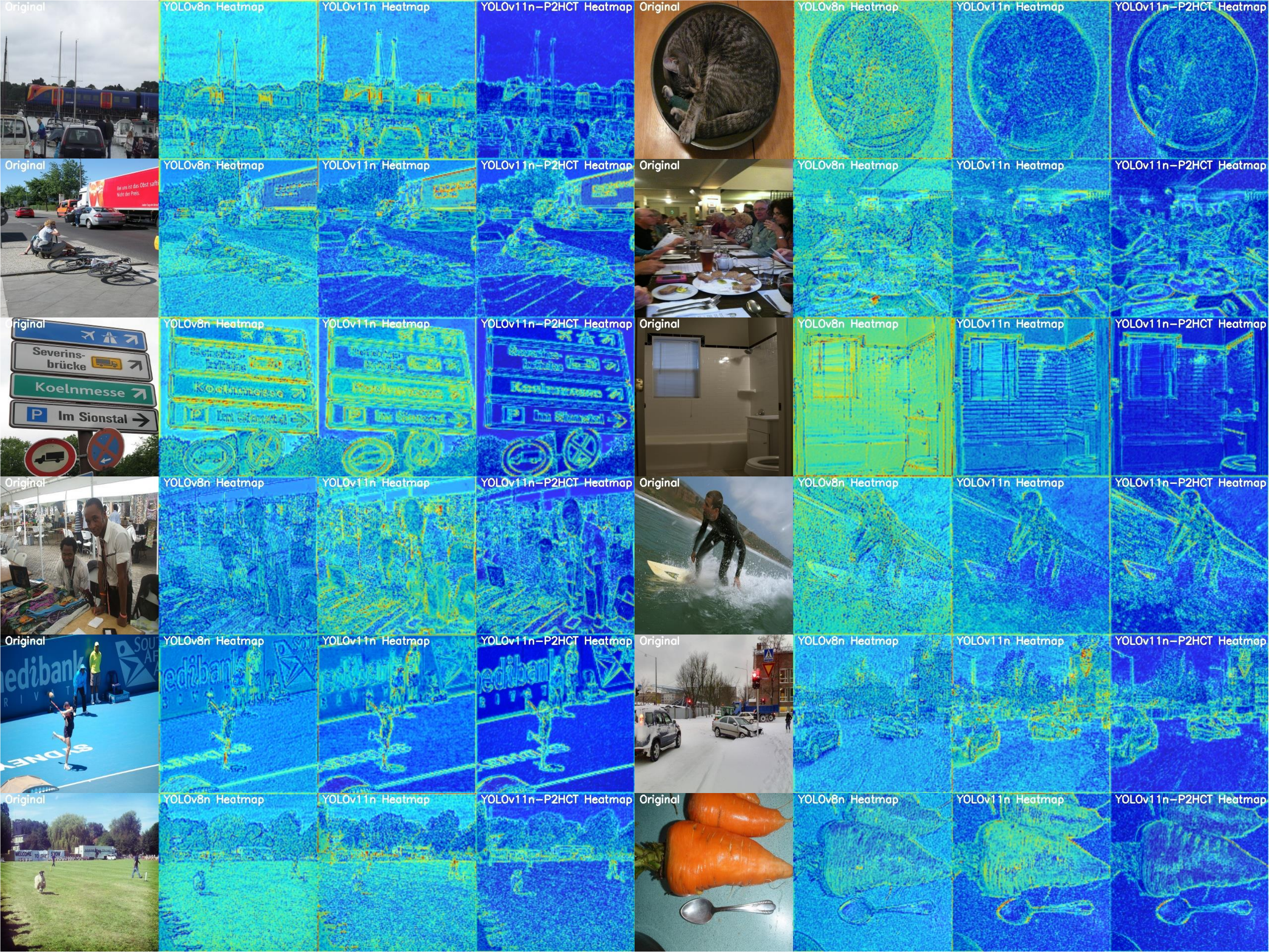}
    \caption{Example attention heatmap results of different methods on COCO 2017 validation set.}
    \label{fig:supp-heatmap}
\end{figure*}

To further illustrate the effectiveness of our P²HCT module, we provide a set of supplementary detection results and corresponding attention heatmaps. Figures \ref{fig:supp-visual} and \ref{fig:supp-heatmap} display qualitative comparisons on challenging scenes—including small, occluded, and densely packed objects—showing both input images with predicted bounding boxes and overlaid fine-attention activation maps. These heatmaps highlight the regions our module focuses on during inference and demonstrate its ability to accurately localize salient features across scales. Altogether, these additional visualizations corroborate the robustness and interpretability of our method under diverse conditions.

\end{document}